%% file: LargeScaleSfM.tex
\documentclass[10pt,twocolumn,letterpaper]{article}

\usepackage{3dv}
\usepackage{times}
\usepackage{epsfig}
\usepackage{graphicx}
\usepackage{amsmath}
\usepackage{amssymb}

\usepackage{bm}
\usepackage{xcolor,colortbl}
\usepackage{soul}
\usepackage{multirow}

\input{ParameterNames.tex}

\usepackage[pagebackref=true,breaklinks=true,letterpaper=true,colorlinks,bookmarks=false]{hyperref}

\threedvfinalcopy 

\def\threedvPaperID{74} 

\newcommand\blfootnote[1]{%
  \begingroup
  \renewcommand\thefootnote{}\footnote{#1}%
  \addtocounter{footnote}{-1}%
  \endgroup
}

\ifthreedvfinal\fi
\begin{document}

\title{GraphMatch: Efficient Large-Scale Graph Construction \\ for Structure from Motion}

\newcommand\Mark[1]{\textsuperscript#1}

\author{\hspace{-3em}Qiaodong Cui\Mark{1}\\ \\
\hspace{-3em} \Mark{1}University of California,\\
\hspace{-3em}Santa Barbara\\
{\tt \small \{qiaodong@umail, psen@ece\}.ucsb.edu}
\and
\hspace{-6em} Victor Fragoso\Mark{2}\\ \\
\and 
\hspace{0em}Chris Sweeney\Mark{3}\\ \\
\hspace{-9em} \Mark{2}West Virginia University\\
\hspace{-9em} {\tt\small victor.fragoso@mail.wvu.edu}
\and \hspace{-3em}Pradeep Sen\Mark{1}\\ \\
\hspace{-2em} \Mark{3}University of Washington\\
\hspace{-2em}  {\tt\small csweeney@cs.washington.edu}
}

\maketitle


\input{Abstract.tex}

\input{Introduction.tex}

\input{PreviousWork.tex}

\input{TheoreticalBackground.tex}

\input{ProposedMethod.tex}


\input{Experiments.tex}


\input{Conclusion.tex}

{\small
\bibliographystyle{ieee}
\bibliography{LargeScaleSfM}
}

\end{document}


\title{GraphMatch: Efficient Large-Scale Graph Construction \\ for Structure from Motion \\ Supplemental Material}

\newcommand\Mark[1]{\textsuperscript#1}

\author{\hspace{-3em}Qiaodong Cui\Mark{1}\\ \\
\hspace{-3em} \Mark{1}University of California,\\
\hspace{-3em}Santa Barbara\\
{\tt \small \{qiaodong@umail, psen@ece\}.ucsb.edu}
\and
\hspace{-6em} Victor Fragoso\Mark{2}\\ \\
\and 
\hspace{0em}Chris Sweeney\Mark{3}\\ \\
\hspace{-9em} \Mark{2}West Virginia University\\
\hspace{-9em} {\tt\small victor.fragoso@mail.wvu.edu}
\and \hspace{-3em}Pradeep Sen\Mark{1}\\ \\
\hspace{-2em} \Mark{3}University of Washington\\
\hspace{-2em}  {\tt\small csweeney@cs.washington.edu}
}

\maketitle


\input{IntroductionSuppl.tex}
\input{Implementation.tex}

\input{ExperimentsForSupplemental.tex}

\input{Pseudo-Code/PseudoCode.tex}
\input{VisualizationsSuppl.tex}

{\small
\bibliographystyle{ieee}
\bibliography{LargeScaleSfM}
}

%% file: ParameterNames.tex
\newcommand{\MaxNumNeighbors}{{\sc MaxNumNeighbors}}
\newcommand{\NumPropagatedPerNeighbor}{{\sc NumToPropagate}}
\newcommand{\MaxTestsPerImage}{{\sc MaxTestsForSampling}}
\newcommand{\NumberSampleIterations}{{\sc NumberSampleIterations}}

%% file: Abstract.tex
\begin{abstract}
\vspace{-0.1in}
\noindent We present {\em GraphMatch}, an approximate yet efficient method for building the matching graph for large-scale structure-from-motion~(SfM) pipelines. Unlike modern SfM pipelines that use vocabulary (Voc.) trees to quickly build the matching graph and avoid a costly brute-force search of matching image pairs, GraphMatch does not require an expensive offline pre-processing phase to construct a Voc.\ tree. Instead, GraphMatch leverages two priors that can predict which image pairs are likely to match, thereby making the matching process for SfM much more efficient. The first is a score computed from the distance between the Fisher vectors of any two images. The second prior is based on the graph distance between vertices in the underlying matching graph. GraphMatch combines these two priors into an iterative ``sample-and-propagate'' scheme similar to the PatchMatch algorithm. Its sampling stage uses Fisher similarity priors to guide the search for matching image pairs, while its propagation stage explores neighbors of matched pairs to find new ones with a high image similarity score. Our experiments show that GraphMatch finds the most image pairs as compared to competing, approximate methods while at the same time being the most efficient.
\end{abstract}

%

%% file: Introduction.tex

\vspace{-6mm}
\section{Introduction}
\label{sec:Introduction}

\blfootnote{Published at IEEE 3DV 2017} Recently, structure-from-motion (SfM) algorithms have achieved impressive reconstructions from large photo collections~\cite{agarwal2009building, frahm2010building, heinly2015reconstructing, schonberger2016structure, snavely2008modeling}. These exciting results are possible thanks to recent improvements in bundle adjustment~\cite{agarwal2010bundle}, motion/camera parameter estimation~\cite{bujnak20093d, chatterjee2013efficient, crandall2013sfm, fragoso2013evsac, fragoso2017ansac, govindu2006robustness, hartley2011l1, jiang2013global, kanatani2000closed, martinec2007robust, moulon2013global, ozyesil2015robust, sweeney2015optimizing, wilson2014robust}, and feature matching~\cite{cheng2014fast, muja2014scalable}.

All modern large-scale SfM pipelines~\cite{agarwal2011building, moulon2013global, schonberger2016structure} require the building of a {\em matching graph} in order to reconstruct a scene from a large photo collection. In this graph, images are nodes and edges represent matching image pairs that share visual content that could be used for 3D-reconstruction. Despite the aforementioned advancements, however, the problem of finding high-quality, matching image pairs to form the matching graph remains a major bottleneck for large-scale SfM pipelines. The reason for this is that the vast amount (75 - 95\%) of image pairs do not match in most photo collections~\cite{wu2013towards}. 

An effective, but prohibitively expensive, approach to filter non-matching image pairs is the brute force method: an exhaustive test of all possible $\mathcal{O}(N^2)$ image pairs. To accelerate the search for suitable image pairs to match, several state-of-the-art large-scale SfM pipelines use image-retrieval techniques~\cite{chum2007total, nister2006scalable, oliva2001modeling}, assuming that image pairs with high visual similarity scores are likely to match. Among the adopted image retrieval techniques, the most implemented in publicly available large-scale SfM pipelines~\cite{agarwal2011building, moulon2013global, schonberger2016structure} is Vocabulary (Voc.) Trees~\cite{nister2006scalable}. 

\begin{figure*}[t]
    \centering
    \includegraphics[width=\textwidth]{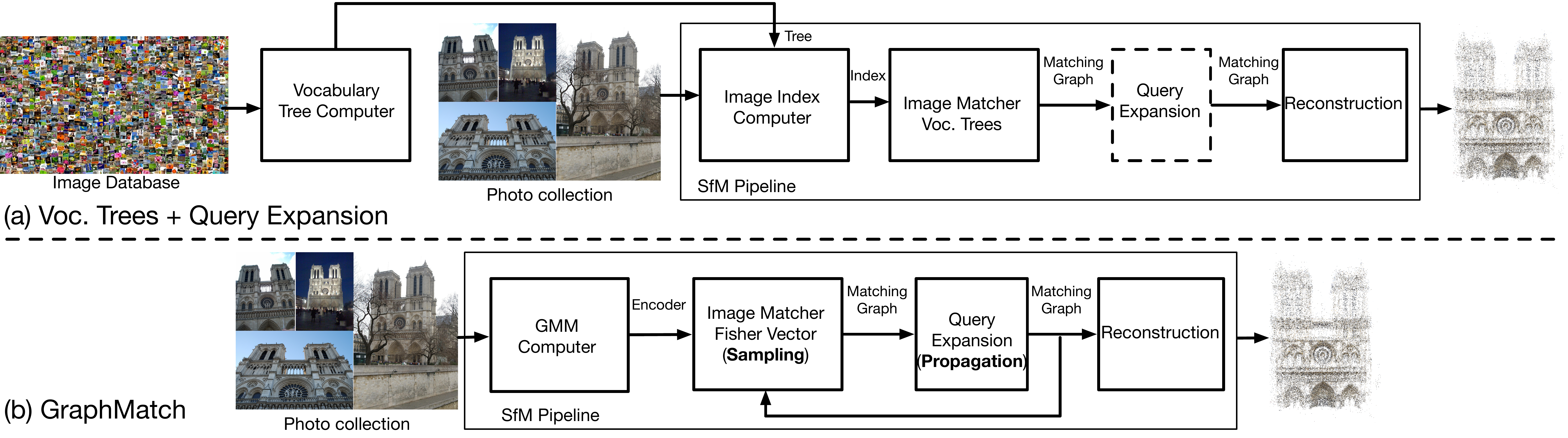}
    \caption{{\bf (a)} Modern, state-of-the-art SfM pipelines, such as the Building-Rome-in-a-Day (BRIAD) pipeline~\cite{agarwal2011building}, accelerate the search for potential image pairs using Vocabulary (Voc.) Trees. However, this requires an expensive offline stage that clusters a large database of image features to construct the Voc.\ Tree, and also has to build an index of the images which consume a large amount of memory. Furthermore, finding potential image matching pairs with Voc.\ Trees can produce a non-dense matching graph that can affect reconstruction quality. To reduce this effect, pipelines such as BRIAD use a {\em query-expansion} step, which uses the current matching graph to identify new potential pairs. {\bf (b)} On the other hand, our proposed GraphMatch approach does not use Voc.\ Trees, and hence eliminates this expensive offline stage. Instead, GraphMatch uses Fisher distances~\cite{perronnin2007fisher, perronnin2010improving} to measure image similarity, since we find that they are better priors for finding matching pairs than Voc.\ trees similarity scores. GraphMatch then follows an iterative ``sample-and-propagate'' scheme, where the sampling process finds suitable image pairs to match and the propagation step uses the current state of the matching graph and image similarities to discover new matching pairs.
}\label{fig:overview}
\end{figure*}

Although Voc.\ Trees help SfM pipelines find suitable image pairs to match more quickly, constructing a Voc.\ Tree is computationally expensive (due to the computation of a tree encoding visual words and the image index) and can demand a large memory footprint~\cite{philbin2007object, stewenius2012size}. Specifically, creating the Voc.\ Tree requires an expensive clustering pass (\eg, k-means~\cite{lloyd1982least}) on a large set or subset of local image features (\eg, SIFT~\cite{lowe2004distinctive}) for every node in the tree. For this reason, SfM pipelines construct this tree in an independent process before executing the reconstruction pipeline. In addition, creating an image index requires passing each of the local features of every image in a dataset through the vocabulary tree, which is also computationally expensive. Another disadvantage of Voc.\ trees is that SfM pipelines have to load the image index in memory to find suitable image pairs to match in a large photo collection. In this scenario, substantial memory footprints are necessary since the image index is large as well. SfM pipelines can reduce the memory footprint by creating a coarse Voc.\ tree. However, a coarse Voc.\ tree typically decreases matching performance.



Since Voc.\ Trees were devised for image-retrieval applications and not for SfM, they only find a fraction of the potentially matching pairs (see Sec.~\ref{sec:experiments}), which decreases reconstruction quality.  
To increase the number of potentially matching image pairs computed by the Voc.\ trees, some SfM pipelines such as the Building-Rome-in-a-Day (BRIAD) pipeline~\cite{agarwal2011building} execute a {\em query-expansion} step after building an initial matching graph (see Fig.~\ref{fig:overview}a).  This idea, drawn from work on text/document retrieval~\cite{chum2007queryexpansion}, is based on the observation that  if  images $i$ and $j$ are matched, and image $j$ is also matched with image $k$, then potentially image $i$ could be matched with $k$.  

In BRIAD,  potentially-matching image pairs are found by applying query expansion a repeated number of times.  This approach has a fundamental problem, however, in that repeated query expansions lead to {\em drift}~\cite{agarwal2011building}, where the context of the images found rapidly diverge from that of the initial image, especially as it starts examining the neighbors of the neighbors, and so on.  Although the geometric verification step in the matching process ensures that these pairs are not admitted into the graph as edges, it greatly reduces the efficiency of the overall matching process because it tests more pairs unlikely to share edges.  In other words, in approaches like BRIAD, the more it performs query expansion, the less efficient the algorithm becomes.

To combat these problems, in this work we present a new algorithm called {\it GraphMatch} (see Fig.~\ref{fig:overview}b), an efficient, approximate matching algorithm that significantly accelerates the search for image pairs in SfM pipelines. GraphMatch is inspired by the PatchMatch algorithm~\cite{Barnes2009}, which has been used successfully for accelerating the search for patch-wise correspondences between two images. Compared to algorithms such as BRIAD, GraphMatch has two key differences which greatly improve its performance.

First, GraphMatch does not use Voc.\ trees, so it does not require an offline expensive stage to construct them or the image index. Instead, we use a score computed from the distance between Fisher vectors~\cite{perronnin2007fisher, perronnin2010improving} of any two images. Our experiments demonstrate that Fisher vector distances are faster to compute and more indicative of a possible match than Voc.\ Tree similarity scores, or those of other descriptors such as VLAD~\cite{jegou2010aggregating} or GIST~\cite{oliva2001modeling}. 

Second, we use an alternative approach to the query expansion step that is based on the PatchMatch algorithm~\cite{Barnes2009}. This alternative approach maximizes the fraction of matching pairs and ensures high-quality reconstruction. PatchMatch makes a similar observation to that of query expansion: if patch $i$ in one image and patch $j$ in another are matched, then the neighboring patches of patch $j$ are likely to be matches for patch $i$. This information is used by PatchMatch in an iterative ``sample-and-propagate'' scheme to efficiently compute the correspondence fields between patches of two images.  First, the correspondences of each patch are initialized to point to random patches in the sampling step. Then in the propagation step, the valid correspondences found in the previous step are propagated to their neighbors. This algorithm iterates until the full, approximate correspondence field has been found.

GraphMatch implements a similar iterative ``sample-and-propagate'' algorithm which builds the matching graph incrementally. However, rather than doing random sampling as in PatchMatch, the sampling stage of GraphMatch uses Fisher scores to search for matching image pairs more effectively. The propagation stage then uses the current graph of matched image pairs to explore the neighbors of images belonging to geometrically verified image pairs. Unlike the query expansion in BRIAD which is only executed after the graph has been constructed and therefore suffers from drift, GraphMatch alternates sampling and propagation in an iterative fashion to find suitable image pairs.


The ``sample-and-propagate'' scheme of GraphMatch provides an excellent balance between matching images with similar visual appearance (the sample stage) while also taking advantage of local connectivity cues (the propagate stage). As a result, GraphMatch is able to efficiently determine which image pairs are likely to produce a good match, avoiding unnecessary overhead in attempting to match image pairs which are unlikely to match.

%% file: PreviousWork.tex
\section{Previous Work}
\label{sec:PreviousWork}

Determining \textit{correspondence} between pairs of images is a critical first key step in recovering scene geometry through SfM. Typical SfM pipelines first detect point features such as SIFT~\cite{lowe1999object} features, then\ perform feature matching between SIFT descriptors in pairs of images to estimate their epipolar geometry. Pairs of images which produce a sufficient number of feature correspondences that pass an additional geometric verification step are then used during SfM to generate feature tracks. Given an input dataset of $N$ images that we wish to reconstruct, brute-force matching approach exhaustively tests all possible image pairs leading to an $O(N^2)$ algorithm. However, most image pairs do not lead to successful matches. Large-scale SfM pipelines can gain efficiency by leveraging information that helps the pipeline identify image pairs that are more likely to match and filter image pairs that are unlikely to match. Wu~\cite{wu2013towards} measures this likelihood by matching only a subset of features, observing that an image pair is likely to match when these subset-features produce a sufficient number of image correspondences. While this strategy provides a significant increase in efficiency, it can yield to disconnected reconstructions~\cite{wu2013towards} and is still requires testing every image pair.

To mitigate the cost of exhaustive matching approaches, image retrieval techniques have been employed to efficiently determine a set of candidate image pairs for feature matching based on image similarity. Vocabulary (Voc.) trees~\cite{nister2006scalable} are frequently used to efficiently match datasets. A vocabulary is learned (typically from clustering image features), and the visual words are organized into a tree that may be efficiently searched. Term Frequency Inverse Document Frequency (TF-IDF) is used to determine the similarity of images using inverted files~\cite{nister2006scalable}. Voc. Trees are highly scalable due to efficient search of the tree structure. Additional image retrieval works have used a learning-based approach to predict which images contain visual overlap. Sch\"{o}nberger~\etal~\cite{schonberger2015paige} uses a machine-learning-based approach dubbed PAIGE to predict which images contain visual overlap. PAIGE trains a random forest (RF)~\cite{ho1998random} using image-pair features to predict the pairs with scene overlap. 

Image retrieval techniques offer efficient performance but typically are optimized based on precision and recall, not for the final SfM reconstruction quality. As such, these methods are suboptimal for use with SfM. Shen~\etal~\cite{shen2016graph} proposed a graph-based approach for SfM image matching where image pairs to match are chosen based on a combination of appearance similarity and local match graph connectivity. After an initial minimum spanning tree of matches is built, the minimum spanning tree is expanded to form strongly consistent loops (determined from pairwise geometry between images in the loops) in the match graph. Finally, community detection is used to densify the matching graph efficiently given the match graph structure after strongly consistent loops are added.

Our approach, GraphMatch, is of similar spirit to the method of Shen~\etal because it also considers visual similarity and match graph connectivity priors. Shen's algorithm, however, aims to find the fewest number of matches possible that provide a strong loop consistency in the underlying matching graph while still yielding high quality reconstructions. Thus, they attempt to find a minimal stable set of edges that yield high quality reconstructions. GraphMatch, on the other hand, aims to accelerate the search for valid matches to be as efficient as possible (by minimizing the number of ``bad'' image pairs tested) so that as many good matches may be found as efficiently as possible. By finding as many valid matches as possible, GraphMatch is able to avoid disconnected reconstructions that plague alternative methods due to weak or missing connectivity~\cite{wu2013towards} and recover more cameras in the resulting SfM reconstruction (\cf Fig.~\ref{fig:NumCameraDatasets}). Indeed, our algorithm typically is able to recover 80-90\% of the total good matches found by the baseline in just a fraction of the time. Further, GraphMatch uses simple-yet-effective priors that yield an easy-to-implement and efficient algorithm.

%% file: TheoreticalBackground.tex

\section{Finding Image Pairs with Informative Priors}
\label{sec:theory}

The task of finding all geometrically verified, matching image pairs in a large database of $N$ images can be posed as the problem of discovering a matching graph. In this graph, vertices represent images, and edges exist when pairs of images contain a sufficient number of geometrically verified feature matches.

\begin{figure}[t]
\centering
	\includegraphics[width=\linewidth]{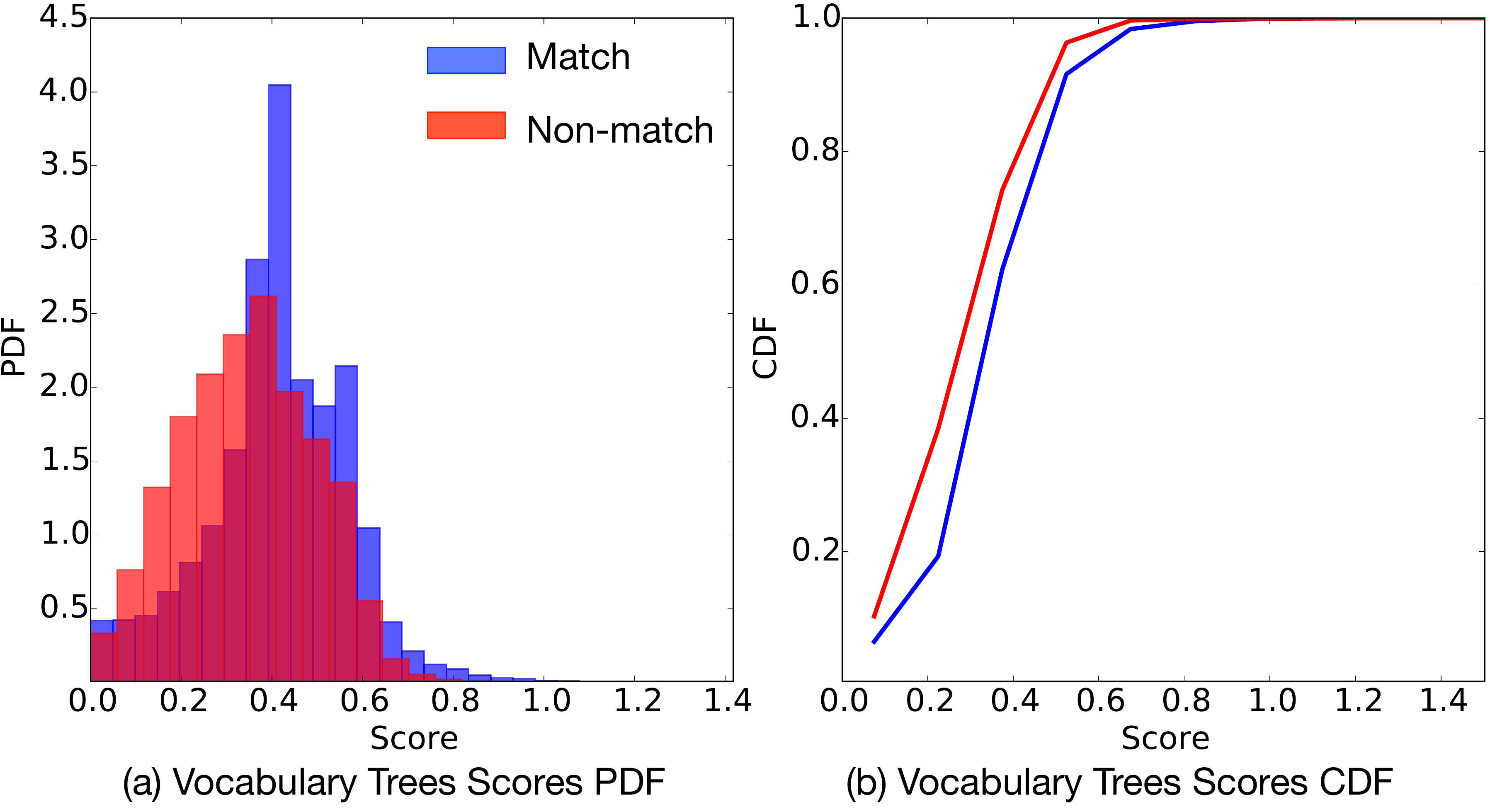}
\caption{\label{fig:VocTreePerformance} Probability density functions (PDFs) and cumulative distribution functions (CDFs) of geometrically-verified matching image pairs (edges) and non-matching image pairs (non-existing edges) as a function of Vocabulary Tree scores. The support of the PDFs and CDFs of valid and non-existing edges overlap significantly.
}
\end{figure}

The brute force approach leads to poor performance for large $N$ because of its $O(N^2)$ complexity. However, it is possible to obtain high quality reconstructions by using {\em approximate matching algorithms} that only use subsets of potential edges~\cite{Snavely2008}. These approximate methods are extremely useful for large-scale SfM applications when they discover a large percentage of edges quickly.

The complexity of finding suitable image pairs with approximate methods depend on the density of the matching graph. The lower the density (or higher the sparsity), the more challenging for algorithms to find image pairs. To alleviate this issue, existing SfM pipelines use Voc.\ trees as a way to generate image pairs with a high similarity score. Unfortunately, the image representation that Voc.\ trees use provides limited information about matching image pairs. This is confirmed by observing the distributions of the Voc.\ trees similarity scores for matching and non-matching image pairs of several large-scale photo collections: {\sc NYC Library, Montreal Notre Dame, Madrid Metropolis, Alamo, Tower of London, Roman Forum}. We can observe in Fig.~\ref{fig:VocTreePerformance} that the distributions of Voc.\ trees similarity scores for matching and non-matching image pairs overlap significantly. The larger the separation between these distributions, the more useful information to find matching image pairs.

\begin{figure*}[t]
\centering
\includegraphics[width=\textwidth]{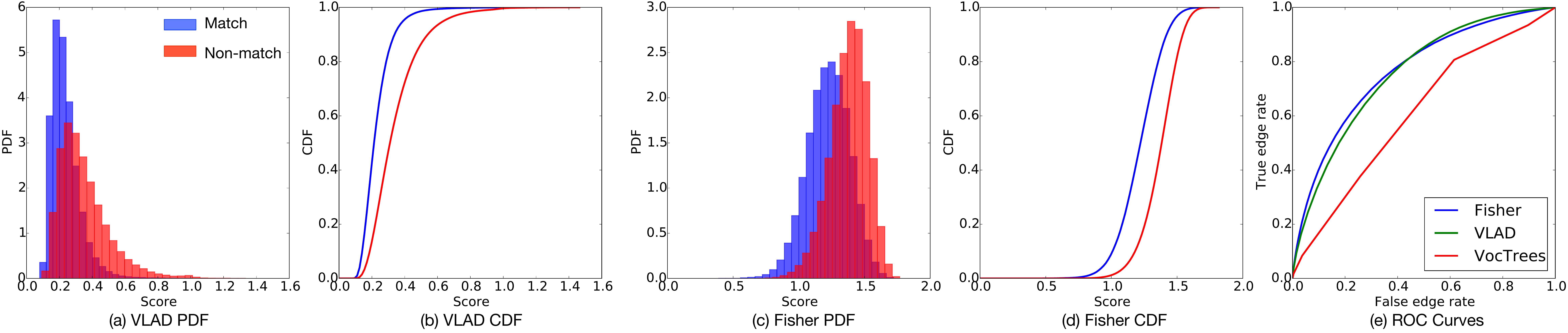}
\caption{\label{fig:VLADvsFisher} PDFs ({\bf (a)} and {\bf (c)}) and CDFs ({\bf (b)} and {\bf (d)}) of edges and non-edges over scores of VLAD ({\bf(a, b)}) and Fisher ({\bf(c, d)}). The overlap between the PDFs of the edges and non-existing edges corresponding to Fisher and VLAD scores is less than that of the Voc.\ Trees scores. The  ROC curves shown in Fig.\ {\bf (e)} confirm this. The curves corresponding to Fisher (blue) and VLAD (green) are above of the curve corresponding to Voc.\ Trees scores (red). Also, Fig.\ {\bf (e)} shows that Fisher scores tend to be better than VLAD scores for predicting edges when the false edge rate is less tha 0.4. Consequently, Fisher scores are the best prior for edge prediction considering a low false edge rate.
\vspace{-0.1in}
}
\end{figure*}

For our algorithm, we begin exploring better priors for edge-finding by studying metrics for global image similarity.  In particular, we tested VLAD~\cite{jegou2010aggregating} and Fisher vectors~\cite{perronnin2007fisher, perronnin2010improving}. As Fig.~\ref{fig:VLADvsFisher} shows, we found that Fisher scores gave us the best separation between the distributions of edges and non-edges, meaning that they provide better guidance towards sampling edges. While the fisher vector is effective, other priors like PAIGE \cite{schonberger2015paige} may be used.

For our second prior, we were inspired by query-expansion~\cite{chum2007queryexpansion} and the PatchMatch algorithm~\cite{Barnes2009}, which has been very successful in image processing tasks~\cite{Barnes2010,Darabi2012}. PatchMatch accelerates the process of finding matching image patches with a clever, two-stage algorithm. First, it randomly samples the correspondence field by testing each patch in the first image with a {\em random} patch in the source. Then, the propagation step will exploit the few good matches that the sampling step found to generate new correspondences. The basic idea of propagation is that neighboring patches in the first image usually correspond to neighboring patches in the second image. Therefore, this stage propagates good matches throughout the image. The algorithm then iterates, repeating the sampling and propagation stages until the approximate correspondence field between the two images is quickly discovered.

\begin{figure}[t]
\centering
\includegraphics[width=\linewidth]{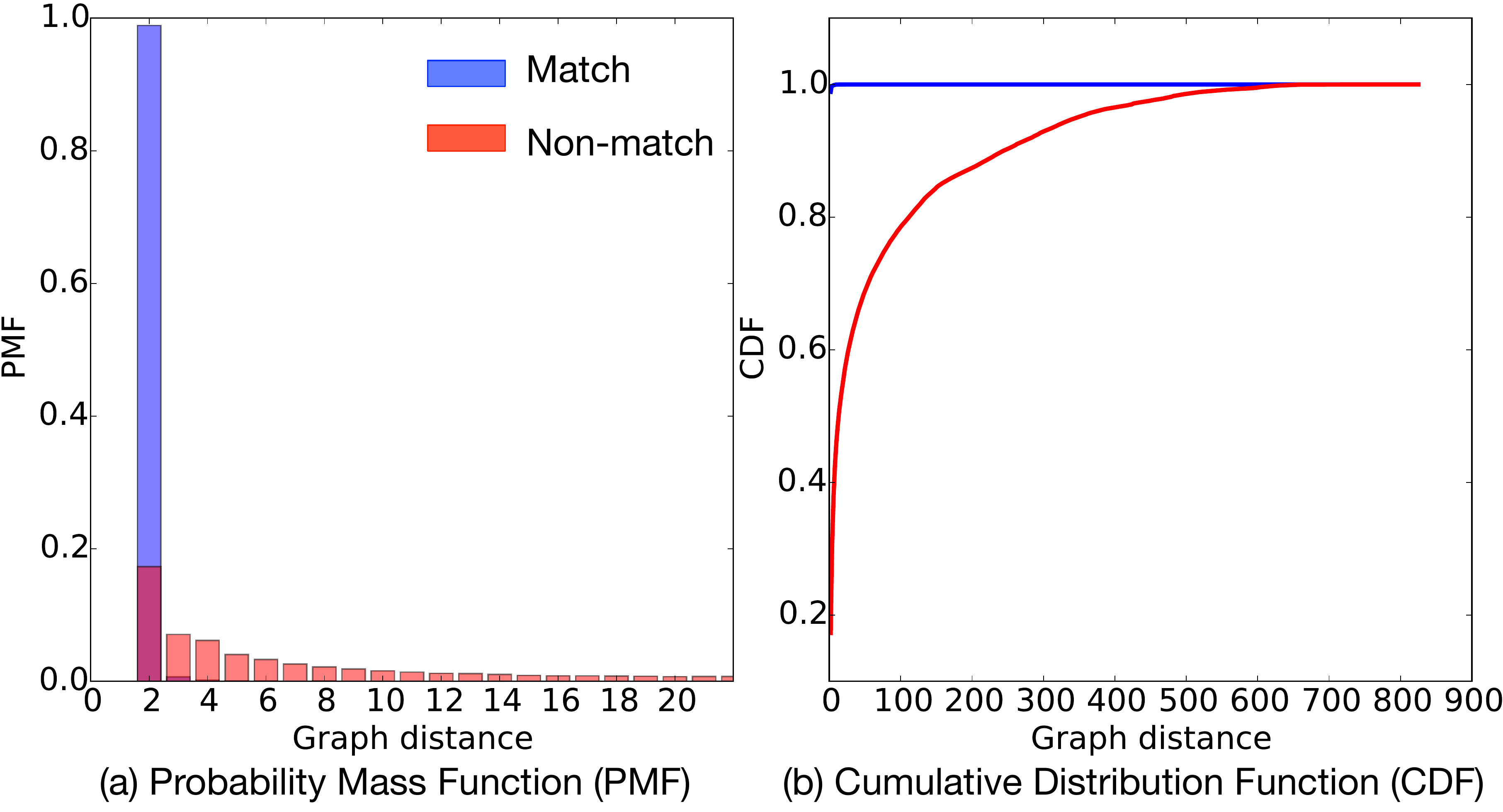}
\caption{\label{fig:GraphDistance} Probability Mass Functions (PMFs) and CDFs for edges and non-edges as a function of a modified graph distance between every pair of vertices $(A, B)$ in the graph. If no edge exists betwen $A$ and $B$, we simply use their normal graph distance (number of edges in their shortest path).  But if an edge exists, we remove it before computing their graph distance (otherwise it would always be `1').  The PMF plot {\bf (a)} shows that $(A,B)$ are much more likely to share an edge when the distance is $2$ (i.e, they have a neighbor in common), which constitutes the foundation of the propagation step in GraphMatch. The CDFs plot {\bf (b)} shows that non-matching pairs have a higher modified graph distance since its CDF (red curve) grows slower than that of the matching pairs (blue curve).
\vspace{-0.1in}
}
\end{figure}

In this work, if two vertices $A$ and $B$ are neighbors (\ie, are connected by an edge in the graph), we hypothesize that the neighbors of $B$ might also be neighbors for $A$ and vice-versa. Indeed, in practice we see that this is the case (see Fig.~\ref{fig:GraphDistance}), where we show the distributions for edges and non-edges as a function of the modified graph distance between the vertices. This fact makes sense for graphs in structure-from-motion applications, where vertices that share an edge usually represent images that are captured in close spatially proximity, and therefore tend to have commonality with other images that are also in close proximity because of spatial coherence. 


Given these two priors, we propose to use them in an alternating algorithm we call {\em GraphMatch}. Unlike existing SfM pipelines that execute a sampling-like step using Voc.\ trees followed by a query-expansion step, GraphMatch iteratively executes a sampling-and-propagation steps until the current matching graph satisfies certain criteria.

%% file: ProposedMethod.tex

\section{Proposed Algorithm: GraphMatch}


Like PatchMatch, GraphMatch has two main steps: {\em sampling} and {\em propagation}. The purpose of its sampling stage is to find previously unexplored matching pairs, which will connect regions of the graph that had not been connected before. Unlike PatchMatch, rather than exploring completely at random, we guide our search towards candidate edges that are more likely to be valid by using the Fisher-distance prior discussed in the last section. The propagation step in GraphMatch aligns well with a query-expansion step and serves a similar purpose to the one in PatchMatch. The goal of the propagation step is to exploit current matching image pairs to discover new matching pairs. 

There are some fundamental differences between PatchMatch and GraphMatch. For example, in PatchMatch the neighborhood of each patch is clearly defined by the parametrization of the underlying pixel grid of an image. However, in the graph discovery case, it is not clear what the ``neighborhood'' of a vertex is, especially since the graph has not been fully discovered. Therefore, inspired by the query-expansion, GraphMatch uses all the current, direct neighbors of a vertex (\ie, the set of vertices that currently share a valid edge with the vertex) as the potential neighborhood of candidates that will be tested for more matches. Unlike query-expansion, however, GraphMatch identifies the candidate image pairs to test, ranks them based on their Fisher distances, and selects a subset as the pairs to geometrically verify. Thus, the propagation step in GraphMatch combines cues from the current state of the matching graph and image similarity priors.

Another fundamental difference between GraphMatch and PatchMatch, is that PatchMatch finds a single correspondence for a given query patch. In contrast, in the graph discovery case, each image can have multiple matching images. Thus, GraphMatch generalizes PatchMatch in this sense by adjusting the sampling and propagation algorithms to test multiple images in the same iteration, and allow each image to have multiple matches.

In the following subsections, we describe the different stages of our algorithm in detail, which takes in as input a large collection of images $\mathbf{I} = \{I_1,\ldots,I_N\}$ and outputs an approximate connectivity graph $\mathbf{G}$.  Complete pseudocode can be found in the supplementary material.

\input{Preprocessing}

\input{SamplingStep.tex}


\input{PropagationStep.tex}


%% file: Preprocessing.tex

\subsection{Pre-processing}
\label{sec:Preprocessing}

The GraphMatch algorithm first pre-computes SIFT features~\cite{lowe2004distinctive} for all $N$ images in the collection to be used later on for the geometric verification process. Unlike existing SfM pipelines that use Voc.\ trees to represent images and measure image similarity, GraphMatch uses Fisher vectors~\cite{perronnin2007fisher, perronnin2010improving}. It uses them for the following reasons: 1) outperform Voc.\ trees-image-based representations for image retrieval~\cite{perronnin2010large}; 2) require only a single clustering pass instead of several passes as in Voc.\ trees; 3) reveal more discriminative information about matching image pairs than Voc.\ trees as shown in Fig.~\ref{fig:VocTreePerformance}.

A fundamental difference between Voc.\ trees-based image matchers and GraphMatch is that Voc. trees-based methods compute the tree using a large database of image features (\eg, SIFT) and several clustering passes in an offline stage. In contrast, GraphMatch only uses the input photo-collection to compute image representations via Fisher vectors as part of the same SfM pipeline. Thus, after computing the local image features of the input photo collection, GraphMatch proceeds to compute a Fisher vector for every image. To do so, GraphMatch first constructs a database of image features by taking a random sample of features for every image in the photo collection. Then, it estimates the cluster priors, diagonal-covariance matrices, and centroids that parametrize the Gaussian Mixture Model (GMM)~\cite{murphy2012machine}. Note that the GMM parameter estimation phase is the only clustering pass to the data in GraphMatch. Using the estimated GMM, GraphMatch computes a Fisher vector for every image in the input photo collection using a Fisher-vector encoder. GraphMatch uses the efficient and multi-threaded GMM estimator and Fisher encoder included in the VlFeat~\cite{vedaldi2010vlfeat} library.


Once every image has its Fisher vector, GraphMatch computes a distance matrix between every pair of images. To compute this matrix efficiently given its $O(N^2)$ complexity, GraphMatch exploits the symmetry in the matrix (\ie, $d(A, B) = d(B, A)$), avoids computing the distance of an image with itself, and computes the matrix in parallel. Consequently, GraphMatch ends up computing half the matrix quickly by using multi-threaded schemes. The distance estimation only involves subtracting two vectors of 4096 floats and taking the dot product of the result with itself. We observed in our experiments that this matrix evaluation takes at most 3\% of the total matching time, and it is not a bottleneck in our approach. The clustering approach in ~\cite{frahm2010building} can be leveraged to further decrease the complexity of this step.


Finally, the last step of the pre-processing stage uses the Fisher distances to create a list for every image that ranks all other images based on proximity to the given image. In our case, images at the beginning of the list have smaller Fisher distances (are closer or similar) to the image in question. GraphMatch implements this ranking step also using efficient multi-threaded schemes. Once the algorithm has completed the pre-processing step, it begins the main body of the algorithm, which iterates between a sampling step and a propagation step and continues until no sampling and propagation has occurred.

%% file: SamplingStep.tex

\subsection{Sampling Step}
\label{sec:Sampling}

The sampling step attempts to find connections between new regions in the graph by testing new potential edges.  In the original PatchMatch algorithm this testing was done at random, but in our case we find that the Fisher distance prior (see Fig.~\ref{fig:VLADvsFisher}) helps guide the sampling and improves the efficiency of the algorithm by increasing the probability that edges are found. 

In order to sample based on Fisher distance, GraphMatch uses the ranked lists for every image that were pre-computed in the pre-processing step and pulls a fixed number of candidate images from each list at every iteration to be used for sampling. The number of images to be pulled is controlled by two parameters: \MaxTestsPerImage, which controls the maximum number of times GraphMatch can test an image before it stops sampling altogether, and \NumberSampleIterations, which governs the maximum number of iterations for which it does sampling. The number of samples for each image per iteration is simply set by the division of these two numbers.

Note that GraphMatch only samples from a vertex if it has less than {\MaxNumNeighbors}, which is part of our termination criteria.  However, this vertex can still be chosen to sample to (\eg, if it is high on another vertices sampling list).  Once the list of sampling pairs has been computed for all vertices in the graph, GraphMatch passes it to a function that will test each one for edges using geometric verification. Those that are found to have edges are added to the graph and to a data structure that tracks all neighboring vertices for each image. Furthermore, the vertices tested are removed from the ranked lists of the appropriate vertices so they are not tested again.


%% file: PropagationStep.tex
\subsection{Propagation Step}
\label{sec:Propagation}

The goal of the propagation step is to identify new edges by leveraging the spatial coherence normally found in the matching graphs for SfM and image similarities computed via Fisher vectors. The propagation step loops over every pair of vertices with known edges in graph $\mathbf{G}$.  Given a pair $A$ and $B$ that share an edge, the propagation step takes the top {\it ranked} neighbors of $B$ (based on their Fisher distance) and tests them against $A$ and vice-versa.  Note that GraphMatch propagates only from vertices that have less than {\MaxNumNeighbors} for our termination criterion.  The number of vertices selected to propagate from each neighbor is given by the parameter {\NumPropagatedPerNeighbor}.

%% file: Experiments.tex

\input{MasterTable.tex}


\begin{figure}
\centering
\includegraphics[width=1\linewidth]{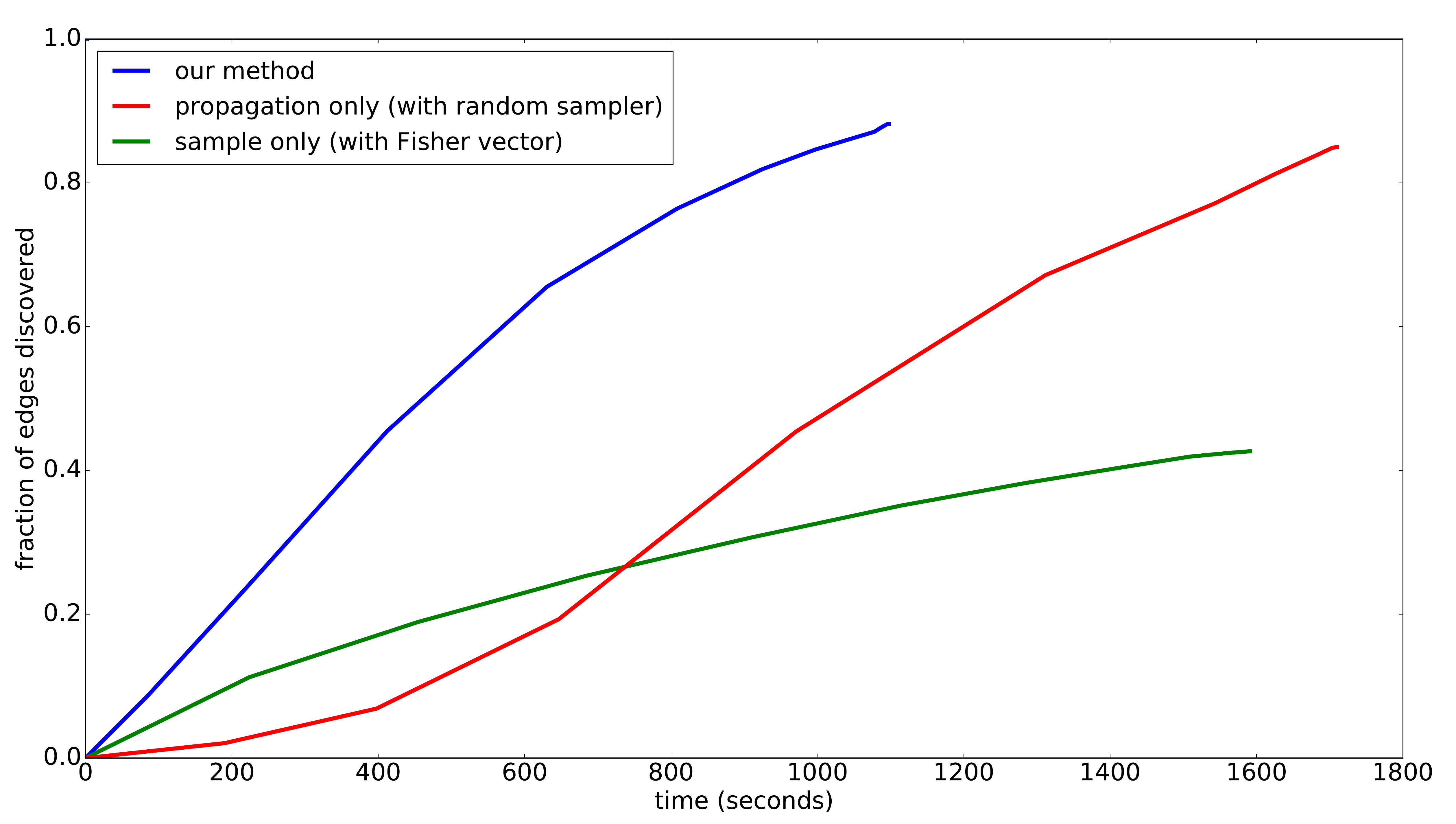}
\caption{The fraction of discovered edges for GraphMatch with different configurations: 1) GraphMatch with sample and propagation stages on (blue); 2) GraphMatch with only the sampling stage (green); and 3) GraphMatch with only propagation stage (red). GraphMatch with both stages on outperforms the other two configurations. GraphMatch leverages the information from both stages to discover edges quickly.}
\label{fig:stage_contribution}
\vspace{-4mm}
\end{figure}

\begin{figure}[t]
\centering
\includegraphics[width=\linewidth]{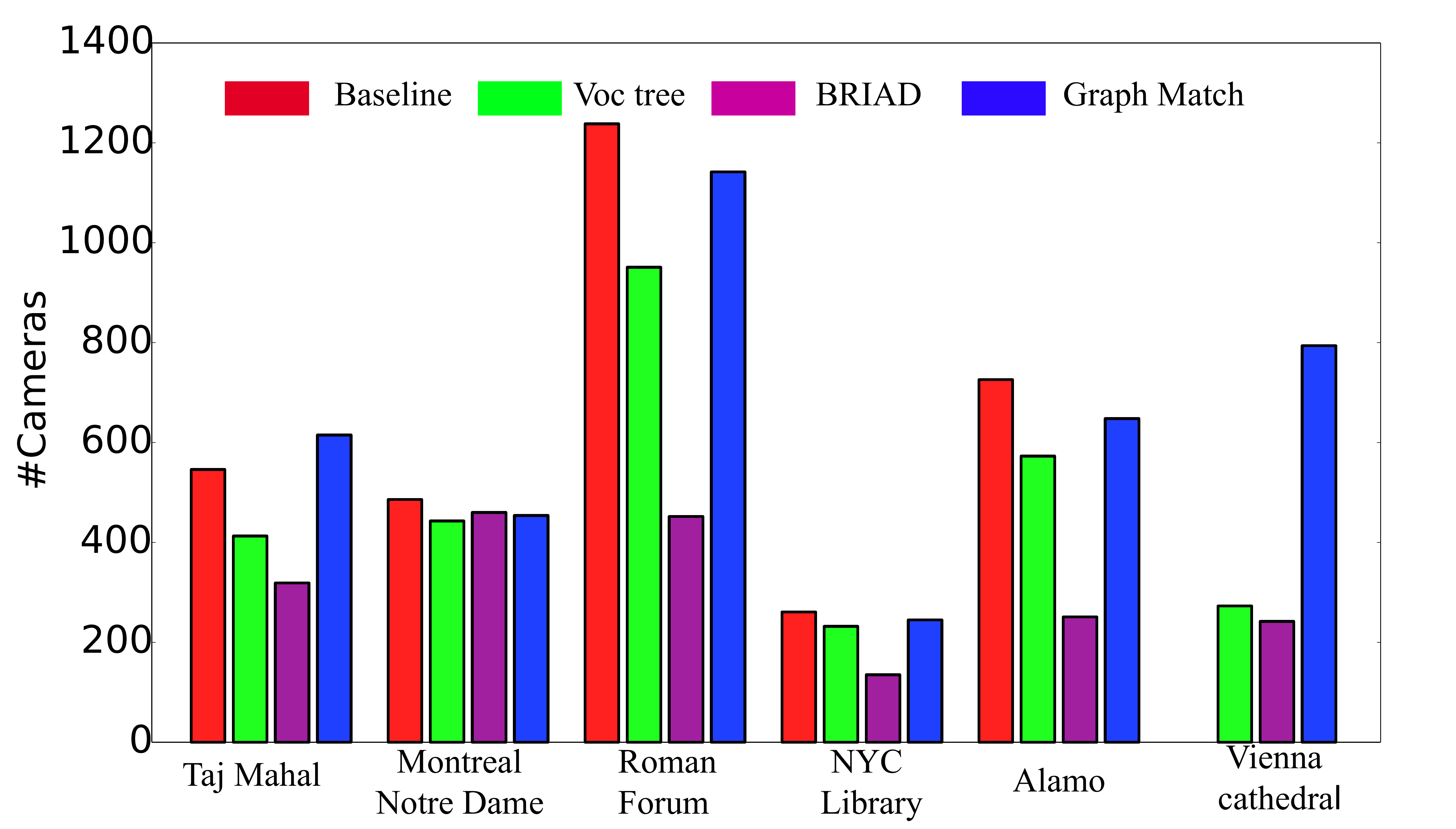}
\caption{\label{fig:NumCameraDatasets}
Number of cameras reconstructed for each method for all scenes in Table~\ref{tab:MasterTable}. GraphMatch consistently reconstructs most cameras and is very close to baseline.}
\vspace{-4mm}
\end{figure}


\section{Experiments}
\label{sec:experiments}

We implemented GraphMatch in C++ using the Theia SfM~\cite{sweeney2015theia} library's API, and incorporated it into Theia's reconstruction pipeline. All experiments were run on a machine with 128 GB of RAM, 2.6 TB of storage space using SSDs, and 2$\times$ Intel Xeon at 2.60GHz each with 8 cores. Since both our implementation and Theia library use multithreading, we launched our SfM pipeline using 32 threads.

\begin{figure*}[t!]
\centering
	\includegraphics[width=\linewidth]{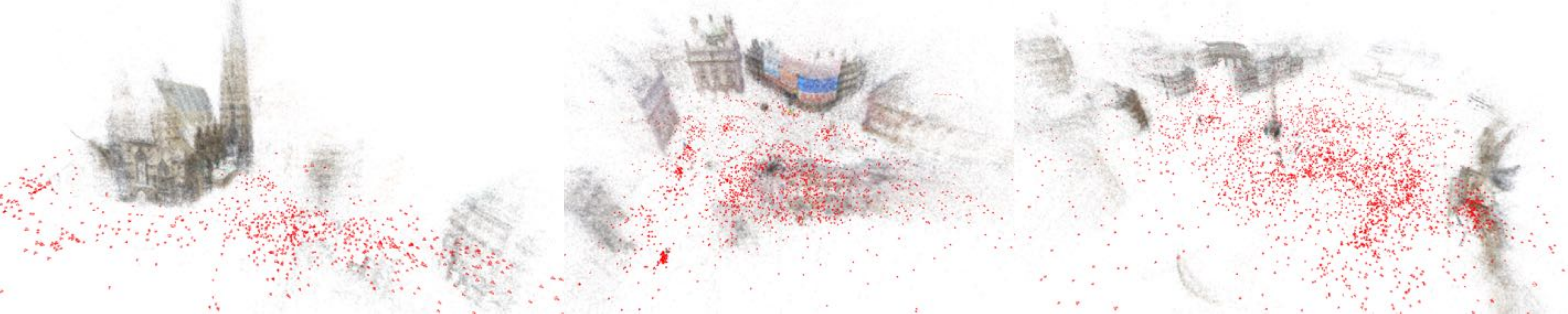}
\caption{\label{fig:LargeSceneReconstructions}Reconstructions of three large datasets, from left to right: {\sc Vienna Cathedral} (6,288 images), {\sc Picadilly} (7,351 images), {\sc Trafalgar} (15,685 images).}
\vspace{-4mm}
\end{figure*}

The experiments were performed using datasets formed from internet photo collections obtained from crawling Flickr~\cite{wilson2014robust}. These photo collections pose interesting challenges to the SfM pipelines because their size ranges from $1497 - 15685$ and the density of the underlying matching graph (\ie, the number of good matches out of all possible image pairs) ranges from $0.04$ to less than $0.004$. As discussed in Section~\ref{sec:theory}, the density determines the chances for an algorithm to find image pairs. The lower the density, the more challenging to find image pairs, and vice-versa.


We compare our approach to the brute-force method (the baseline), the commonly used Voc.\ trees method, and the building Rome in a day (BRIAD) matching scheme which implements a Voc.\ trees followed by a query-expansion. We compare their timings for pre-processing, matching, and total reconstruction. In addition, we measure the reconstruction quality using the discovered matching graph by computing the number of recovered cameras, number of edges in the matching graph (\ie, good matches), and in most cases the mean and median distances of the cameras positions to their corresponding camera positions computed with the baseline method. For the baseline method, we used the multi-threaded brute-force (exhaustive search) implementation provided by Theia. For Voc.\ Tree-based methods (plain Voc.\ trees and BRIAD), we used a pre-computed and publicly available tree~\footnote{http://people.inf.ethz.ch/jschoenb/colmap/} with 1 million visual words, and indexed each dataset using the multi-threaded COLMAP SfM~\cite{schonberger2016structure} library. Also, BRIAD retrieved $k=40$ NN for each image while Voc.\ tree method use $k=120$. The timings for the pre-processing phase of Voc.\ trees-based methods only consider the time for creating the image index, since we use a pre-computed tree. In BRIAD, we executed four times the query-expansion as suggested by Agrawal~\etal~\cite{agarwal2009building}. We did not consider the pre-emptive method by Wu~\cite{wu2013towards}, since Sch\"{o}nenberger~\etal~\cite{schonberger2015paige} reported that Voc.\ trees outperform such a pre-emptive method.

The results of each method can be seen in Table~\ref{tab:MasterTable}. Among the approximate methods (\ie, BRIAD and Voc.\ trees), our algorithm achieves the highest reconstruction quality. This is because it recovers the highest number of cameras and the highest number of edges in the matching graph; see \#recon-cameras and \#edges columns and Fig.~\ref{fig:NumCameraDatasets}. Our algorithm recovers a great number of cameras and is the most comparable to that of the baseline. Thus, GraphMatch is able to consistently find more edges than Voc. Tree-based methods because our algorithm is much more efficient than Voc.\ trees at finding good image pairs to match. This leads to more stable SfM reconstructions with longer feature tracks and better visual coverage~\cite{sweeney2015optimizing}. This can be seen in the Taj Mahal dataset, where the Voc.\ Tree method recovers very few edges and is only able to reconstruct 68\% of the number of cameras with respect to the baseline. See supp. material for results on more datasets.

We also computed the efficiency of the four algorithms as the ratio between the number of found matching image pairs and the total number of tested image pairs. The average efficiency of the baseline was 0.021, which is exactly the same as the average density for all scenes as expected. The average efficiency for Voc.Trees-based methods was slightly higher at 0.08, and ours was 0.27 (nearly $10\times$ more than baseline). Furthermore, we broke down the efficiency by stage for our algorithm. We found that the sampling step was 0.08 efficient, while the propagation step was 0.37 efficient. The fact that nearly one third of the edges tested by propagation are valid given that the average density of all the scenes is around 0.021 underscores the importance of our propagation step. To understand the effect of iteratively alternating sampling and propagation steps, we measured the fraction of edges discovered over time with both stages alternating each other, with only sampling stage on and with only propagation stage on. As shown in fig.~\ref{fig:stage_contribution}, our alternating scheme discovers edges better. Thus, both steps complement each other to boost the edge discovery.


To study whether a trained Voc.\ trees per datasets could improve its performance, we trained Voc.\ trees for the dataset {\sc Taj Mahal} and {\sc Roman Forum}. We found that the matching efficiency and reconstruction quality remained basically the same. It reconstructed 422 and 975 cameras respectively, while the tree from COLMAP reconstructed 413 and 951 cameras. The training time for these two datasets was 0.9 and 1.5 hours, respectively. This brings the total pre-processing time to 1.1 and 1.8 hours, respectively, while fisher vectors required only 0.9 and 1.4 minutes to train for the same datasets, respectively; an approximately 70x speedup compared to Voc. Trees. This suggests that training the Voc. Tree on specific scenes is not a good option for large-scale SfM, especially when compared to the proposed approach.



To assess the reconstruction quality of large-scale photo collections using GraphMatch, we reconstruct three additional datasets: {\sc Vienna Cathedral} (6288 images), {\sc Picadilly} (7531 images), and {\sc Trafalgar} (15685 images). The SfM pipeline with our algorithm required the following reconstruction timings: {\sc Picadilly} in 9.7 hours (estimated 163.6 hours with the baseline) and {\sc Trafalgar} in 16.0 hours (estimated 835.4 hours with the baseline). For more results and analysis please see the supp. material.



%% file: MasterTable.tex

\begin{table*}[t]
\centering
\scriptsize{
  \begin{tabular}{|c|c|c||c|c|c||c|c||c|c|c||c||c|c|}
    \hline
    \multirow{2}{*}{Dataset} & \# images & \multirow{2}{*}{Alg.} & {\# recon } & \multirow{2}{*}{\# edges} & \multicolumn{2}{c||}{Position Error (m)} & \multicolumn{4}{c||}{Time (min)} & \multicolumn{2}{c|}{Speed Up} \\
    \cline{2-2}\cline{6-7}\cline{8-13}
    & graph density &  & cameras & & avg. & median & pre & match & recon & total & match & overall\\
    \hline
     \hline
     \multirow{4}{*}{{\sc Taj Mahal}} & \multirow{2}{*}{1497}  &
     baseline & 576 & 63,474 & - & - & 0 & 696.62 & 38.81 & 754.62 & 1$\times$ & 1$\times$\\
     \cline{3-7}\cline{8-13}
     & & voc. tree & 413 & 9,368 & 0.04 & 0.03 & 10.04 & 37.39 & 53.60 & 109.23 & 12.80$\times$ & 6.91$\times$\\
     \cline{2-2} \cline{3-7}\cline{8-13}
     & \multirow{2}{*}{0.0437} & BRIAD  & 319 & 21,332 & 0.04 & 0.03 & 10.04 & 26.98 & 13.14 & 58.19 & 15.79$\times$ & 12.97 $\times$\\
     \cline{3-7}\cline{8-13}
     & & ours & 615 & 48,949 & \textbf{0.03}  & \textbf{0.02} & 0.87 & 41.78 & 57.62 & 108.54 & 14.02$\times$ & 6.95 $\times$\\
     \hline
%
%
     \hline
     \multirow{4}{*}{{\sc Montreal N.D.}} & \multirow{2}{*}{2298} & 
     baseline & 486 & 33,836 & - & - & 0 & 1556.23 & 51.30 & 1643.39 & 1$\times$ & 1$\times$\\
     \cline{3-7}\cline{8-13}
     & & voc. tree & 431 & 18,060 & 0.08 & 0.07 & 18.35 & 56.66 & 37.16 & 123.04 & 18.44 $\times$ & 13.36 $\times$\\
     \cline{2-2}\cline{3-7}\cline{8-13}
     & \multirow{2}{*}{0.0100} & BRIAD  & 430 & 8,241 & \textbf{0.06} & \textbf{0.03} & 18.35 & 32.51 & 52.50 & 113.82 & 25.77$\times$ & 14.44 $\times$\\
     \cline{3-7}\cline{8-13}
     & & ours & 460 & 31968 & 0.06 & 0.05 & 1.39 & 48.90 & 48.62 & 109.65 & 26.01 $\times$ & 14.99 $\times$\\
     \hline
%
     \hline
     \multirow{4}{*}{{\sc Roman Forum}} & \multirow{2}{*}{2364} &
     baseline & 1247 & 52,155 & - & - & 0 & 1839.29 & 58.78 & 1939.45 & 1$\times$ & 1$\times$\\
     \cline{3-7}\cline{8-13}
     & & voc. tree & 951 & 15,795 & 0.67 & 0.56 & 20.48 & 91.88 & 71.50 & 196.11 & 15.03$\times$ & 9.89 $\times$\\
     \cline{2-2} \cline{3-7}\cline{8-13}
     &  \multirow{2}{*}{0.0159} & BRIAD  & 452 & 26,730 & \textbf{0.03} & \textbf{0.02} & 20.48 & 61.47 & 35.68 & 129.37 & 19.96 $\times$ & 14.99 $\times$\\
     \cline{3-7}\cline{8-13}
     & & ours & 1142 & 50,216 & 0.12  & 0.09 & 1.39 & 78.80 & 52.81 & 145.04 & 20.35 $\times$ & 13.37 $\times$\\
     \hline
%
     \hline
     \multirow{4}{*}{{\sc NYC Library}} & \multirow{2}{*}{2550} &  
     baseline & 261 & 15,241 & - & - & 0 & 1065.78 & 24.98 & 1119.26 & 1$\times$ & 1$\times$\\
     \cline{3-7}\cline{8-13}
     & & voc. tree & 232 & 7,639 & \textbf{3.82} & \textbf{2.28} & 20.18 & 56.94 & 18.37 & 107.18 & 12.25$\times$ & 10.44 $\times$\\
     \cline{2-2}\cline{3-7}\cline{8-13}
     & \multirow{2}{*}{0.0039} & BRIAD  & 135 & 4,327 & 4.18 & 2.76 & 20.18 & 39.44 & 9.02 & 80.03 & 15.30$\times$ & 13.99 $\times$\\
     \cline{3-7}\cline{8-13}
     & & ours & 245 & 13,427 & 4.21 & 3.11 & 1.40 & 41.46 & 24.53 & 78.82 & 20.09$\times$ & 16.67 $\times$\\
     \hline
%
     \hline
     \multirow{4}{*}{{\sc Alamo}} & \multirow{2}{*}{2915} & 
     baseline & 726 & 62,793 & - & - & 0 & 2506.32 & 85.31 & 2646.70 & 1$\times$ & 1$\times$\\
     \cline{3-7}\cline{8-13}
     & & voc. tree & 573 & 19,932 & 0.21 & 0.08 & 27.64  & 73.37 & 44.93 & 160.47 & 22.04$\times$ & 16.49 $\times$\\
      \cline{2-2}\cline{3-7}\cline{8-13}
     & \multirow{2}{*}{0.0122} & BRIAD  & 251 & 12,490 & 0.20 & 0.10 &  27.64 & 41.28 & 6.33 & 89.25 & 30.64 $\times$ & 29.66 $\times$\\
     \cline{3-7}\cline{8-13}
     & & ours & 648 & 50,943 & 0.13  & 0.04 & 1.61 & 61.29 & 65.82 & 143.04 & 33.07 $\times$ & 18.50 $\times$ \\
     \hline

     \hline
     \multirow{3}{*}{{\sc Vienna Cathedral}} & \multirow{2}{*}{6288} & 
     voc. tree & 273 & 10,578 & \textbf{-} & \textbf{-} & 117.36  & 450.80 & 20.05 & 624.96 & - & -\\
     \cline{3-7}\cline{8-13}
     & & BRIAD  & 242 & 17,578 & - & - &  117.36 & 216.60 & 22.94 & 389.75 & - & - \\
     \cline{2-2}\cline{3-7}\cline{8-13}
     &  0.004 & ours & 794 & 79,394 & \textbf{-}  & \textbf{-} & 3.37 & 367.58 & 44.28 & 450.60 & -& -\\
     \hline

  \end{tabular}}
  \caption{Results of timing and camera positions errors for different algorithms on different scenes. Note the preprocessing time does not include time to extract the sift features.}
  \label{tab:MasterTable}
  \vspace{-4mm}
\end{table*}

%% file: Conclusion.tex

\section{Conclusion}
\label{sec:Conclusion}
In this paper, we have presented a novel algorithm called {\em GraphMatch} for image matching in structure-from-motion pipelines that use scene priors to efficiently find high quality image matches. It iteratively searches for image matches with a ``sample-and-propagate'' strategy similar to that of PatchMatch~\cite{Barnes2009, Barnes2010}. During the sampling stage, we use priors computed from Fisher vector~\cite{perronnin2007fisher, perronnin2010improving} distances between images to guide the search for image pairs which are likely to match. Inspired by the query-expansion step~\cite{agarwal2009building}, the propagation stage exploits local connectivity and image similarities to find new matches based on neighbors of the current matches. This strategy is able to efficiently discover which image pairs are most likely to yield good matches without requiring an offline training process. Our experiments show that GraphMatch achieves the highest number of recovered cameras and the highest efficiency (\ie, the ratio between the matching image pairs and number of pairs tested), while maintaining equivalent speed-ups compared to that of Voc.\ tree-based methods without considering the time of the tree construction. Consequently, GraphMatch becomes an excellent algorithm for discovering the matching graph efficiently in large-scale SfM pipelines without requiring an expensive offline stage for building a Voc.\ tree.


While our method is effective, it has a few limitations: 
\begin{enumerate}
\item Optimal parameter tuning. Our method has four parameters that were optimized over some datasets, and we have not found theoretically the best parameters for each datasets; and
\item Image representation to guide the sampling stage. GraphMatch uses Fisher vectors and are effective. However, there may be other methods more suitable for this task (\eg, PAIGE \cite{schonberger2015paige}).
\end{enumerate}
Nevertheless, GraphMatch is effective for efficient image matching in structure-from-motion pipelines.

\vspace{-4mm}
\paragraph{Acknowledgments.} This work was supported in part by NSF grants IIS-1321168, IIS-1342931, IIS-1619376 and IIS-1657179. The authors would like to thank Atieh Taheri, who worked on an early version of the algorithm and conducted some preliminary experiments.

%% file: IntroductionSuppl.tex
\section{Introduction}

In this document we present implementation details in Section~\ref{sec:Implementation}; additional experiments showing performance advantages of GraphMatch, and experiments analyzing reconstruction quality and efficiency in Section~\ref{sec:AdditionalExperiments}; the pseudo-code of the implementation in Section~\ref{sec:pseudocode}; and visualizations of reconstructed photo-collections using the proposed GraphMatch algorithm in Section~\ref{sec:Visualizations}.

%% file: Implementation.tex
\section{Implementation Details}
\label{sec:Implementation}

In this section, we provide complete implementation details to allow for reproducibility of our method, although we will release our code when the paper is published.  The complete pseudocode for our algorithm is given in Algs.~1 -- 4.  
Note that all  pair lists used in the code (e.g., $\mathbf{P}$, $\mathbf{P}_s$,  or $\mathbf{P}_p$) are implemented using unordered sets so duplicate pairs will be removed.

Our method has only four parameters that are fixed for all scenes.  First is {\NumberSampleIterations}, which controls the maximum number of iterations in which sampling will be done.  After this number of iterations, our algorithm will only do propagation.  We found that a value of $10$ provided a reasonable balance between sampling and propagation.  Next, {\MaxNumNeighbors} sets the maximum number of neighbors a vertex can have, after which sampling/propagating from this vertex will stop, although we can still sample/propagate to it from another vertex.  This is one of our termination criteria that will stop the algorithm (otherwise it would continue sampling until all edges are tested) and is set to 120, since many algorithms like vocabulary trees~\cite{nister2006scalable}. 

{\NumPropagatedPerNeighbor} sets how many neighboring vertices from each neighbor we can propagate to and is set to $4$. This means that every neighbor of a given vertex will provide it at most $4$ images for testing.  Finally, {\MaxTestsPerImage} decides how many tests can be done for each image before sampling stops. This is another one of the termination criteria, and is set to be proportional to the number of input images.  The idea is that the larger the dataset, the more we want to allow it to sample to make sure we find the connected regions since the graph is less dense.  By experimenting on a few scenes, we found that a scaling factor of $0.017$ worked well. Furthermore, we wanted to have a lower limit of $10$ and a maximum of $120$, so in the end {\MaxTestsPerImage} $= \max\left(\min\left( \left(0.017 \times N\right),120\right),10\right)$.

Our implementation of GraphMatch followed a multithreaded scheme for sampling and propagation stages. We implemented the sampling stage following a map-reduce~\cite{dean2008mapreduce} approach. First, we divide the vertices to explore into groups or shards and enqueue them in a threadpool. Then, each worker in the threadpool pulls a shard and performs the sampling algorithm as described earlier. When all the workers finish processing all the shards, the implementation unifies the candidate image pairs to explore in a set. This set of image pairs is then passed to a feature matcher for geometric verification. Our implementation computes the number of shards as a function of the number of available threads in the pool.

We implemented the propagation step also following a map-reduce approach. However, in this case the input to the propagator is the graph, specifically a list of edges to propagate. Unlike the sampling stage implementation, the shard for the propagation stage is an edge. Thus, each worker in the threadpool pulls a single edge and executes the propagation algorithm described earlier. When all the workers finish processing all the edges, the implementation unifies the new candidate image pairs to explore in a set. Again, the implementation passes this set to a feature matcher for geometric verification.

To keep the memory footprint controlled in both stages, our implementation used an LRU cache. This cache allowed the implementation to keep the most requested information from the state of the graph in memory and fetch the least requested from disk.

To compute the Fisher and VLAD vectors for every image, we used the VlFeat~\cite{vedaldi2010vlfeat} computer vision library. We used 16 centroids for Fisher and VLAD encoders. We set the normalization flags to get improved Fisher vectors~\cite{perronnin2010improving} and used the signed squared-root normalization method~\cite{vedaldi2010vlfeat}. Also, we  used 1,000 features for every image to compute these global descriptors. We calculate the Fisher and VLAD distance matrix in parallel described in the pre-processing stage.

%% file: ExperimentsForSupplemental.tex

\section{Additional Experiments}
\label{sec:AdditionalExperiments}

In this section we show additional experiments that show the advantages of GraphMatch over vocabulary trees, BRIAD, and the baseline method. The focus of these experiments is to understand the performance of GraphMatch by varying different options (\eg, global descriptors), measuring the contribution of each stage in GraphMatch, and measure the reconstruction quality and efficiency of GraphMatch.

\subsection{Performance as a Function of Global Descriptors}

\begin{figure}[t]
\centering
	\includegraphics[width=1\linewidth]{Suppl_Figures/global_descriptors1.pdf}
\caption{The fraction of discovered edges as a function of time and different global descriptors in GraphMatch on Roman Forum. Fisher-vector descriptors (blue) discovers more edges over time than VLAD descriptors (green) curve. Nevertheless, the GIST descriptor (purple) performs comparable for a certain period of time ($< 1500$ sec) and later performs worse than Fisher vectors. GraphMatch outperforms the baseline significantly regardless of the used global descriptor.}
\label{fig:global_descriptors}
\end{figure}
GraphMatch exploits priors based on appearance-based image-similarity metrics and the underlying connectivity of the match graph for a given dataset. First, we explore the effect on the edge discovery as a function of time and different global descriptors.

For this experiment, we consider the following global descriptor scores: Fisher descriptor distances, GIST descriptor distances, and VLAD descriptor distances. These descriptor distances were used to obtain the prior based on image similarities that guides the sampling stage.

Fig.~\ref{fig:global_descriptors} shows the results of this experiment on the Roman Forum dataset. We can observe that Fisher vectors (blue curve) is the method that makes GraphMatch discover the most edges over time. We conclude this is because its fraction of discovered edges is higher than that of VLAD descriptor (green curve), and comparable or better than the curve for GIST (purple). As a final comparison, we observe that any global descriptor used in GraphMatch yields a higher fraction of discovery of edges over time compared to that of the the baseline or bruteforce method (red curve).

\subsection{Contribution of Each Stage In GraphMatch}


\begin{figure}[t]
\centering
	\includegraphics[width=\linewidth]{Figures/fig6_stack.pdf}
\caption{\label{fig:NumEdgesPerIterationBreakdown} Fraction of edges discovered per iteration by the sampling and propagation steps of GraphMatch for the {\sc NYC Library} (bottom) and {\sc Alamo} (top) scenes.}
\label{fig:edge_perstage_iteration}
\end{figure}


The goal of this experiment is to measure the contribution of each stage in GraphMatch over iterations. To this end, we measured the fraction of edges discovered over iterations from sampling stage and propagation stage.


The results of this experiment on the Alamo and NYC library are shown in fig.~\ref{fig:edge_perstage_iteration}. As we can see the sampling step helps at the beginning to connect new regions of the graph, later propagation dominates as the interconnections are computed more densely. The figure shows that the combination of the two stages complement each other, this underlines the importance of alternating the sampling and propagation stage.

\subsection{Reconstruction Quality and Efficiency}


In this section we analyze the reconstruction quality (measured by the number of recovered cameras), and the efficiency of various methods: baseline, Voc.\ trees, BRIAD, and GraphMatch.

\input{EfficiencyTable.tex}

The results of this experiment are summarized in Table~\ref{tab:EfficiencyTable}. This table shows the highest efficiency in bold in the second-to-last column, and the number of cameras that the SfM pipeline recovered from the matching graph in the right-most column. The results of this experiments show that GraphMatch is the most efficient for most datasets at finding edges compared to Voc. Trees and the baseline method accross several scenes. However, GraphMatch tends to be the most efficient for most datasets compared to BRIAD, except on UnionSquare, Piccadilly, and Trafalgar datasets where GraphMatch achieves the second highest efficiency. Nevertheless, in terms of the reconstruction quality, GraphMatch recovers a higher number of edges (or matching image pairs) than all of the approximate methods (\ie, Voc.\ Trees and BRIAD). This positions GraphMatch as a good method to approximate the matching graph computed by the baseline. Thanks to a better approximation than that of the competing, approximate methods, GraphMatch yields the highest number of recovered cameras. As a consequence of the better approximated matching graph, GraphMatch allows SfM pipelines to produce a higher quality reconstruction than that of the competing, approximate methods.






%% file: EfficiencyTable.tex
\begin{table*}[t]
\centering
\caption{Efficiency and reconstruction evaluation for all the tested matching algorithms: baseline, Voc.\ Trees, BRIAD, and GraphMatch. The largest efficiency and number of recovered cameras of each dataset is shown in bold. GraphMatch consistently tends to achieve the highest efficiency in comparison with the baseline, Voc.\ Trees, and BRIAD accross different scenes presenting diverse number of images and graph densities. GraphMatch allows SfM pipelines to recover the highest number of cameras compared to that of the competing, approximate methods. This is because GraphMatch recovers most of the matching image pairs, which makes GraphMatch the method that best approximates the matching graph produced by the baseline.}
\scriptsize{
  \begin{tabular}{|c|c|c|c|c|c|c|c|}
    \hline
    \multirow{2}{*}{Dataset} &  \multirow{2}{*}{\# images} & \multirow{2}{*}{Alg.} &  \# edges &  \# edges &  \multirow{2}{*}{Efficiency}  & \multirow{2}{*}{\# cameras} \\
    &  &  & tested & matched & &   \\
    \hline
  \hline
\multirow{6}{*}{{\sc Taj Mahal}} & \multirow{5}{*}{1497} & 
baseline & 1119756 & 63474 & 0.05668 & 546 \\
\cline{3-7}
& & Voc. Trees & 110996 & 9368 & 0.08439 & 413 \\
\cline{3-7}
& & BRIAD & 102286 & 21332 &  0.20855 & 319 \\
\cline{3-7}
& & Our sampling step & 17486 & 2378 & 0.13599 & \multirow{3}{*}{{\bf 615}} \\
\cline{3-6}
& & Our propagation step & 101799 & 46571 & 0.45747 & \\
\cline{3-6}
& & Our overall algorithm  & 119285 & 48949 &  {\bf 0.41035} & \\
\cline{3-6}
\hline
\multirow{6}{*}{{\sc Montreal Notre Dame}} & \multirow{5}{*}{2298} &
baseline & 2639253 & 33836 & 0.01282 & 486 \\
\cline{3-7}
& & Voc. Trees & 175512 & 18060 & 0.10290 & 431 \\
\cline{3-7}
& & BRIAD & 110343 & 8241 & 0.07468 & 430 \\
\cline{3-7}
& & Our sampling step & 42419 & 4145 & 0.09771 & \multirow{3}{*}{{\bf 460}} \\
\cline{3-6}
& & Our propagation step & 86282 & 27823 & 0.32247 & \\
\cline{3-6}
& & Our overall algorithm & 128701 & 31968 & {\bf 0.24839 } & \\
\cline{3-6}
\hline

\multirow{6}{*}{{\sc Roman Forum}} & \multirow{5}{*}{2364} &
baseline & 2793066 & 52155 & 0.01867 & 1238 \\
\cline{3-7}
& & Voc. Trees & 182358 & 15795 & 0.08661 & 951 \\
\cline{3-7}
& & BRIAD & 197374 & 26730 & 0.13543 & 452 \\
\cline{3-7}
& & Our sampling step & 43886 & 3686 &  0.08399 & \multirow{3}{*}{{\bf 1142}}\\
\cline{3-6}
& & Our propagation step & 152874  & 46530 & 0.30436 & \\
\cline{3-6}
& & Our overall algorithm & 196760 & 50216 & {\bf 0.25521} & \\
\cline{3-6}
\hline

\multirow{6}{*}{{\sc NYC Library}} & \multirow{5}{*}{2550} &
baseline & 3244878 & 15241 & 0.00469 & 261 \\
\cline{3-7}
& & Voc. Trees & 197050 & 7639 & 0.03876 & 232 \\
\cline{3-7}
& & BRIAD & 111135 & 4327 & 0.03893 & 135 \\
\cline{3-7}
& & Our sampling step & 65874 & 2726 & 0.04138 & \multirow{3}{*}{{\bf 245}}\\
\cline{3-6}
& & Our propagation step & 37569 & 10701 & 0.28483 & \\
\cline{3-6}
& & Our overall algorithm & 103443 & 13427 & {\bf 0.12980 } & \\
\cline{3-6}
\hline

\multirow{6}{*}{{\sc Alamo}} & \multirow{5}{*}{2915} &
baseline & 4241328 & 62793 & 0.01480 & 726 \\
\cline{3-7}
& & Voc. Trees & 223907 & 19932 & 0.08902 & 573 \\
\cline{3-7}
& & BRIAD & 142557 & 12490 & 0.08761 & 251 \\
\cline{3-7}
& & Our sampling step & 81484 & 3787 & 0.04648 & \multirow{3}{*}{{\bf 648}} \\
\cline{3-6}
& & Our propagation step & 99510 & 47156 & 0.47388 &  \\
\cline{3-6}
& & Our overall algorithm & 180994 & 50943 & {\bf 0.28146} & \\
\cline{3-6}
\hline

\multirow{5}{*}{{\sc Union Square}} & \multirow{5}{*}{5961} &
Voc. Trees & 611401 & 2580 & 0.00421 & 33 \\
\cline{3-7}
& & BRIAD & 230570 & 11978 & {\bf 0.05195} & 396 \\
\cline{3-7}
& & Our sampling step & 396075 & 4853 & 0.01225 & \multirow{3}{*}{{\bf 478}} \\
\cline{3-6}
& & Our propagation step & 51224 & 12739 & 0.24869 & \\
\cline{3-6}
& & Our overall algorithm & 447299 & 17592 & 0.03933 & \\
\cline{3-6}
\hline

\multirow{5}{*}{{\sc Vienna Cathedral}} & \multirow{5}{*}{6288} &
Voc. Trees & 633504 & 10578 & 0.01670 & 273 \\
\cline{3-7}
& & BRIAD & 281635 & 17578 & 0.06241 & 242 \\
\cline{3-7}
& & Our sampling step & 392353 & 13852 & 0.03530 & \multirow{3}{*}{{\bf 794}}\\
\cline{3-6}
& & Our propagation step & 149515 & 65542 & 0.43836 & \\
\cline{3-6}
& & Our overall algorithm & 541868 & 79394 & {\bf 0.14652} & \\
\cline{3-6}
\hline

\multirow{5}{*}{{\sc Piccadilly}} & \multirow{5}{*}{7351} &
Voc. Trees & 569594 & 28789 & 0.05054 & 1649 \\
\cline{3-7}
& & BRIAD & 436309 & 93785 & {\bf 0.21495} & 1773 \\
\cline{3-7}
& & Our sampling step & 460353 & 14477 & 0.03145 & \multirow{3}{*}{{\bf 1863}} \\
\cline{3-6}
& & Our propagation step & 436226 & 133838 & 0.30681 & \\
\cline{3-6}
& & Our overall algorithm & 896579 & 148315 & 0.16542 & \\
\cline{3-6}
\hline

\multirow{5}{*}{{\sc Trafalgar}} & \multirow{5}{*}{15685} &
Voc. Trees & 1292365 & 46478 & 0.03596 & 1919 \\
\cline{3-7}
& & BRIAD & 947627 & 217338 & {\bf 0.22935} & 2699 \\
\cline{3-7}
& & Our sampling step & 1084254 & 23960 & 0.02298 & \multirow{3}{*}{{\bf 4057}} \\
\cline{3-6}
& & Our propagation step & 875221 & 319300 & 0.36482 & \\
\cline{3-6}
& & Our overall algorithm & 1959475 & 343260 &  0.17518 & \\
\cline{3-6}
\hline

  \end{tabular}}
  \label{tab:EfficiencyTable}
  \vspace{-4mm}
\end{table*}

%% file: VisualizationsSuppl.tex
\section{Visualizations of Reconstructed Scenes}
\label{sec:Visualizations}
In this section we show different visualizations of several reconstructed scenes: Trafalgar~\ref{fig:trafalgar}, Taj Mahal~\ref{fig:taj_mahal}, Mount Rushmore~\ref{fig:mount_rushmore}, Vienna Cathedral~\ref{fig:vienna}, and Piccadilly~\ref{fig:piccadilly}. In addition to the visualizations of these scenes, we also include a video describing GraphMatch, and navigations of three reconstructed scenes: Vienna Cathedral, Piccadilly, and Trafalgar.

\begin{figure*}[t]
\centering
\includegraphics[width=1\linewidth]{Suppl_Figures/trafalgar.png}
\caption{Reconstructed Trafalgar scene using GraphMatch.}
\label{fig:trafalgar}
\end{figure*}

\begin{figure*}
\centering
\includegraphics[width=1\linewidth]{Suppl_Figures/taj_mahal.png}
\caption{Reconstructed Taj Mahal scene using GraphMatch.}
\label{fig:taj_mahal}
\end{figure*}

\begin{figure*}
\centering
\includegraphics[width=1\linewidth]{Suppl_Figures/rushmore_close.png}
\caption{Reconstructed Mount Rushmore scene using GraphMatch.}
\label{fig:mount_rushmore}
\end{figure*}

\begin{figure*}
\centering
\includegraphics[width=1\linewidth]{Suppl_Figures/vienna.png}
\caption{Reconstructed Vienna Cathedral scene using GraphMatch.}
\label{fig:vienna}
\end{figure*}

\begin{figure*}
\centering
\includegraphics[width=1\linewidth]{Suppl_Figures/piccadilly.png}
\caption{Reconstructed Piccadilly scene using GraphMatch.}
\label{fig:piccadilly}
\end{figure*}

%% file: LargeScaleSfM.bbl
\begin{thebibliography}{10}\itemsep=-1pt

\bibitem{agarwal2011building}
S.~Agarwal, Y.~Furukawa, N.~Snavely, I.~Simon, B.~Curless, S.~M. Seitz, and
  R.~Szeliski.
\newblock Building {R}ome in a day.
\newblock {\em Communications of the ACM}, 54(10):105--112, 2011.

\bibitem{agarwal2010bundle}
S.~Agarwal, N.~Snavely, S.~M. Seitz, and R.~Szeliski.
\newblock Bundle adjustment in the large.
\newblock In {\em Proc. of the European conference on computer vision}, 2010.

\bibitem{agarwal2009building}
S.~Agarwal, N.~Snavely, I.~Simon, S.~M. Seitz, and R.~Szeliski.
\newblock Building {R}ome in a day.
\newblock In {\em 2009 IEEE 12th international conference on computer vision},
  pages 72--79. IEEE, 2009.

\bibitem{Barnes2009}
C.~Barnes, E.~Shechtman, A.~Finkelstein, and D.~B. Goldman.
\newblock {P}atch{M}atch: A randomized correspondence algorithm for structural
  image editing.
\newblock {\em ACM Trans. Graph.}, 28(3):24:1--24:11, July 2009.

\bibitem{Barnes2010}
C.~Barnes, E.~Shechtman, D.~B. Goldman, and A.~Finkelstein.
\newblock The generalized {P}atch{M}atch correspondence algorithm.
\newblock In {\em Proceedings of the 11th European Conference on Computer
  Vision Conference on Computer Vision: Part III}, ECCV'10, pages 29--43,
  Berlin, Heidelberg, 2010. Springer-Verlag.

\bibitem{bujnak20093d}
M.~Bujnak, Z.~Kukelova, and T.~Pajdla.
\newblock 3d reconstruction from image collections with a single known focal
  length.
\newblock In {\em 2009 IEEE 12th International Conference on Computer Vision},
  pages 1803--1810. IEEE, 2009.

\bibitem{chatterjee2013efficient}
A.~Chatterjee and V.~Madhav~Govindu.
\newblock Efficient and robust large-scale rotation averaging.
\newblock In {\em Proceedings of the IEEE International Conference on Computer
  Vision}, pages 521--528, 2013.

\bibitem{cheng2014fast}
J.~Cheng, C.~Leng, J.~Wu, H.~Cui, H.~Lu, et~al.
\newblock Fast and accurate image matching with cascade hashing for 3d
  reconstruction.
\newblock In {\em Proc. of the IEEE Conf. on Computer Vision and Pattern
  Recognition}, 2014.

\bibitem{chum2007total}
O.~Chum, J.~Philbin, J.~Sivic, M.~Isard, and A.~Zisserman.
\newblock Total recall: Automatic query expansion with a generative feature
  model for object retrieval.
\newblock In {\em Computer Vision, 2007. ICCV 2007. IEEE 11th International
  Conference on}, pages 1--8. IEEE, 2007.

\bibitem{chum2007queryexpansion}
O.~Chum, J.~Philbin, J.~Sivic, M.~Isard, and A.~Zisserman.
\newblock Total recall: Automatic query expansion with a generative feature
  model for object retrieval.
\newblock In {\em 2007 IEEE 11th International Conference on Computer Vision},
  pages 1--8, Oct 2007.

\bibitem{crandall2013sfm}
D.~J. Crandall, A.~Owens, N.~Snavely, and D.~P. Huttenlocher.
\newblock Sfm with mrfs: Discrete-continuous optimization for large-scale
  structure from motion.
\newblock {\em IEEE transactions on pattern analysis and machine intelligence},
  35(12):2841--2853, 2013.

\bibitem{Darabi2012}
S.~Darabi, E.~Shechtman, C.~Barnes, D.~B. Goldman, and P.~Sen.
\newblock {I}mage {M}elding: Combining inconsistent images using patch-based
  synthesis.
\newblock {\em ACM Trans. Graph.}, 31(4):82:1--82:10, July 2012.

\bibitem{fragoso2013evsac}
V.~Fragoso, P.~Sen, S.~Rodriguez, and M.~Turk.
\newblock {EVSAC: Accelerating Hypotheses Generation by Modeling Matching
  Scores with Extreme Value Theory}.
\newblock In {\em Proc. of the IEEE International Conference on Computer
  Vision}, 2013.

\bibitem{fragoso2017ansac}
V.~Fragoso, C.~Sweeney, P.~Sen, and M.~Turk.
\newblock {ANSAC: Adaptive Non-Minimal Sample and Consensus}.
\newblock In {\em {Proc. of the British Machine Vision Conference}}, 2017.

\bibitem{frahm2010building}
J.-M. Frahm, P.~Fite-Georgel, D.~Gallup, T.~Johnson, R.~Raguram, C.~Wu, Y.-H.
  Jen, E.~Dunn, B.~Clipp, S.~Lazebnik, et~al.
\newblock Building {R}ome on a cloudless day.
\newblock In {\em European Conference on Computer Vision}, pages 368--381.
  Springer, 2010.

\bibitem{govindu2006robustness}
V.~M. Govindu.
\newblock Robustness in motion averaging.
\newblock In {\em Asian Conference on Computer Vision}, pages 457--466.
  Springer, 2006.

\bibitem{hartley2011l1}
R.~Hartley, K.~Aftab, and J.~Trumpf.
\newblock L1 rotation averaging using the weiszfeld algorithm.
\newblock In {\em Computer Vision and Pattern Recognition (CVPR), 2011 IEEE
  Conference on}, pages 3041--3048. IEEE, 2011.

\bibitem{heinly2015reconstructing}
J.~Heinly, J.~L. Schonberger, E.~Dunn, and J.-M. Frahm.
\newblock Reconstructing the world* in six days*(as captured by the yahoo 100
  million image dataset).
\newblock In {\em Proc. of the IEEE Conf. on Computer Vision and Pattern
  Recognition}, 2015.

\bibitem{ho1998random}
T.~K. Ho.
\newblock The random subspace method for constructing decision forests.
\newblock {\em IEEE transactions on pattern analysis and machine intelligence},
  20(8):832--844, 1998.

\bibitem{jegou2010aggregating}
H.~J{\'e}gou, M.~Douze, C.~Schmid, and P.~P{\'e}rez.
\newblock Aggregating local descriptors into a compact image representation.
\newblock In {\em Computer Vision and Pattern Recognition (CVPR), 2010 IEEE
  Conference on}, pages 3304--3311. IEEE, 2010.

\bibitem{jiang2013global}
N.~Jiang, Z.~Cui, and P.~Tan.
\newblock A global linear method for camera pose registration.
\newblock In {\em Proceedings of the IEEE International Conference on Computer
  Vision}, pages 481--488, 2013.

\bibitem{kanatani2000closed}
K.~Kanatani and C.~Matsunaga.
\newblock Closed-form expression for focal lengths from the fundamental matrix.
\newblock In {\em Proc. 4th Asian Conf. Comput. Vision}, volume~1, pages
  128--133. Citeseer, 2000.

\bibitem{lloyd1982least}
S.~Lloyd.
\newblock Least squares quantization in pcm.
\newblock {\em IEEE transactions on information theory}, 28(2):129--137, 1982.

\bibitem{lowe1999object}
D.~G. Lowe.
\newblock Object recognition from local scale-invariant features.
\newblock In {\em Computer vision, 1999. The proceedings of the seventh IEEE
  international conference on}, volume~2, pages 1150--1157. Ieee, 1999.

\bibitem{lowe2004distinctive}
D.~G. Lowe.
\newblock Distinctive image features from scale-invariant keypoints.
\newblock {\em International journal of computer vision}, 60(2):91--110, 2004.

\bibitem{martinec2007robust}
D.~Martinec and T.~Pajdla.
\newblock Robust rotation and translation estimation in multiview
  reconstruction.
\newblock In {\em 2007 IEEE Conference on Computer Vision and Pattern
  Recognition}, pages 1--8. IEEE, 2007.

\bibitem{moulon2013global}
P.~Moulon, P.~Monasse, and R.~Marlet.
\newblock Global fusion of relative motions for robust, accurate and scalable
  structure from motion.
\newblock In {\em Proceedings of the IEEE International Conference on Computer
  Vision}, pages 3248--3255, 2013.

\bibitem{muja2014scalable}
M.~Muja and D.~G. Lowe.
\newblock Scalable nearest neighbor algorithms for high dimensional data.
\newblock {\em IEEE Transactions on Pattern Analysis and Machine Intelligence},
  36(11):2227--2240, 2014.

\bibitem{murphy2012machine}
K.~P. Murphy.
\newblock {\em Machine learning: a probabilistic perspective}.
\newblock MIT press, 2012.

\bibitem{nister2006scalable}
D.~Nister and H.~Stewenius.
\newblock Scalable recognition with a vocabulary tree.
\newblock In {\em 2006 IEEE Computer Society Conference on Computer Vision and
  Pattern Recognition (CVPR'06)}, volume~2, pages 2161--2168. IEEE, 2006.

\bibitem{oliva2001modeling}
A.~Oliva and A.~Torralba.
\newblock Modeling the shape of the scene: A holistic representation of the
  spatial envelope.
\newblock {\em International journal of computer vision}, 42(3):145--175, 2001.

\bibitem{ozyesil2015robust}
O.~Ozyesil and A.~Singer.
\newblock Robust camera location estimation by convex programming.
\newblock In {\em Proc. of the IEEE Conf. on Computer Vision and Pattern
  Recognition}, pages 2674--2683, 2015.

\bibitem{perronnin2007fisher}
F.~Perronnin and C.~Dance.
\newblock {F}isher kernels on visual vocabularies for image categorization.
\newblock In {\em 2007 IEEE Conference on Computer Vision and Pattern
  Recognition}, pages 1--8. IEEE, 2007.

\bibitem{perronnin2010large}
F.~Perronnin, Y.~Liu, J.~S{\'a}nchez, and H.~Poirier.
\newblock Large-scale image retrieval with compressed fisher vectors.
\newblock In {\em Computer Vision and Pattern Recognition (CVPR), 2010 IEEE
  Conference on}, pages 3384--3391. IEEE, 2010.

\bibitem{perronnin2010improving}
F.~Perronnin, J.~S{\'a}nchez, and T.~Mensink.
\newblock Improving the {F}isher kernel for large-scale image classification.
\newblock In {\em European conference on computer vision}, pages 143--156.
  Springer Berlin Heidelberg, 2010.

\bibitem{philbin2007object}
J.~Philbin, O.~Chum, M.~Isard, J.~Sivic, and A.~Zisserman.
\newblock Object retrieval with large vocabularies and fast spatial matching.
\newblock In {\em 2007 IEEE Conference on Computer Vision and Pattern
  Recognition}, pages 1--8. IEEE, 2007.

\bibitem{schonberger2015paige}
J.~L. Schonberger, A.~C. Berg, and J.-M. Frahm.
\newblock Paige: pairwise image geometry encoding for improved efficiency in
  structure-from-motion.
\newblock In {\em Proceedings of the IEEE Conference on Computer Vision and
  Pattern Recognition}, pages 1009--1018, 2015.

\bibitem{schonberger2016structure}
J.~L. Sch{\"o}nberger and J.-M. Frahm.
\newblock Structure-from-motion revisited.
\newblock In {\em Proc. of the IEEE Conference on Computer Vision and Pattern
  Recognition}, 2016.

\bibitem{shen2016graph}
T.~Shen, S.~Zhu, T.~Fang, R.~Zhang, and L.~Quan.
\newblock Graph-based consistent matching for structure-from-motion.
\newblock In {\em European Conference on Computer Vision}, pages 139--155.
  Springer, 2016.

\bibitem{snavely2008modeling}
N.~Snavely, S.~M. Seitz, and R.~Szeliski.
\newblock Modeling the world from internet photo collections.
\newblock {\em International Journal of Computer Vision}, 80(2):189--210, 2008.

\bibitem{Snavely2008}
N.~Snavely, S.~M. Seitz, and R.~Szeliski.
\newblock Skeletal graphs for efficient structure from motion.
\newblock In {\em Proc. of the IEEE Conference on Computer Vision and Pattern
  Recognition}, pages 1--8, jun 2008.

\bibitem{stewenius2012size}
H.~Stew{\'e}nius, S.~H. Gunderson, and J.~Pilet.
\newblock Size matters: exhaustive geometric verification for image retrieval
  accepted for eccv 2012.
\newblock In {\em Computer Vision--ECCV 2012}, pages 674--687. Springer, 2012.

\bibitem{sweeney2015theia}
C.~Sweeney, T.~Hollerer, and M.~Turk.
\newblock Theia: A fast and scalable structure-from-motion library.
\newblock In {\em Proceedings of the 23rd ACM international conference on
  Multimedia}, pages 693--696. ACM, 2015.

\bibitem{sweeney2015optimizing}
C.~Sweeney, T.~Sattler, T.~Hollerer, M.~Turk, and M.~Pollefeys.
\newblock Optimizing the viewing graph for structure-from-motion.
\newblock In {\em Proceedings of the IEEE International Conference on Computer
  Vision}, pages 801--809, 2015.

\bibitem{vedaldi2010vlfeat}
A.~Vedaldi and B.~Fulkerson.
\newblock Vlfeat: An open and portable library of computer vision algorithms.
\newblock In {\em Proceedings of the 18th ACM international conference on
  Multimedia}, pages 1469--1472. ACM, 2010.

\bibitem{wilson2014robust}
K.~Wilson and N.~Snavely.
\newblock Robust global translations with 1{DSfM}.
\newblock In {\em Proc. of the European Conference on Computer Vision}, 2014.

\bibitem{wu2013towards}
C.~Wu.
\newblock Towards linear-time incremental structure from motion.
\newblock In {\em 2013 International Conference on 3D Vision-3DV 2013}, pages
  127--134. IEEE, 2013.

\end{thebibliography}
